\documentclass[runningheads]{llncs}
\usepackage{graphicx}

\usepackage{tikz}
\usepackage{comment} 
\usepackage{amsmath,amssymb} 
\usepackage{color}
\usepackage{multirow}

\usepackage{hyperref}
\hypersetup{hidelinks,
	colorlinks=true,
	allcolors=black,
	pdfstartview=Fit,
	breaklinks=true}

\begin{document}
\pagestyle{headings}
\mainmatter
\def\ECCVSubNumber{25}  

\titlerunning{Do Not Disturb Me} 
\authorrunning{Zhao et al.} 

\title{Do Not Disturb Me: Person Re-identification Under the Interference of Other Pedestrians} 



\author{Shizhen Zhao$^1$\thanks{This work was done when Shizhen Zhao was an intern at Tencent Youtu Lab.}, Changxin Gao$^1$\thanks{Corresponding author.}, Jun Zhang$^2$, Hao Cheng$^2$, Chuchu Han$^1$, Xinyang Jiang$^2$, Xiaowei Guo$^2$, Wei-Shi Zheng$^3$, Nong Sang$^1$, Xing Sun$^2$}

\institute{\textsuperscript{\rm 1}Key Laboratory of Image Processing and Intelligent Control, School of Artificial Intelligence and Automation, Huazhong University of Science and Technology,\\ \textsuperscript{\rm 2}Tencent Youtu Lab, \textsuperscript{\rm 3}Sun Yat-sen University\\
{\tt\small Email: \{zhaosz, cgao\}@hust.edu.cn}
}

\maketitle

\begin{abstract}
In the conventional person Re-ID setting, it is widely assumed that cropped person images are for each individual.
However, in a crowded scene, off-shelf-detectors may generate bounding boxes involving multiple people, where the large proportion of background pedestrians or human occlusion exists. 
The representation extracted from such cropped images, which contain both the target and the interference pedestrians, might include distractive information.
This will lead to wrong retrieval results.
To address this problem, this paper presents a novel deep network termed Pedestrian-Interference Suppression Network (PISNet).
PISNet leverages a Query-Guided Attention Block (QGAB) to enhance the feature of the target in the gallery, under the guidance of the query.
Furthermore, the involving Guidance Reversed Attention Module and the Multi-Person Separation Loss promote QGAB to suppress the interference of other pedestrians. 
Our method is evaluated on two new pedestrian-interference datasets and the results show that the proposed method performs favorably against existing Re-ID methods. 
Our project is available at \href{https://github.com/X-BrainLab/PI-ReID}{https://github.com/X-BrainLab/PI-ReID}.


\end{abstract}

\section{Introduction}

\begin{figure*}[!tb]
\begin{center}
\includegraphics[width=8cm]{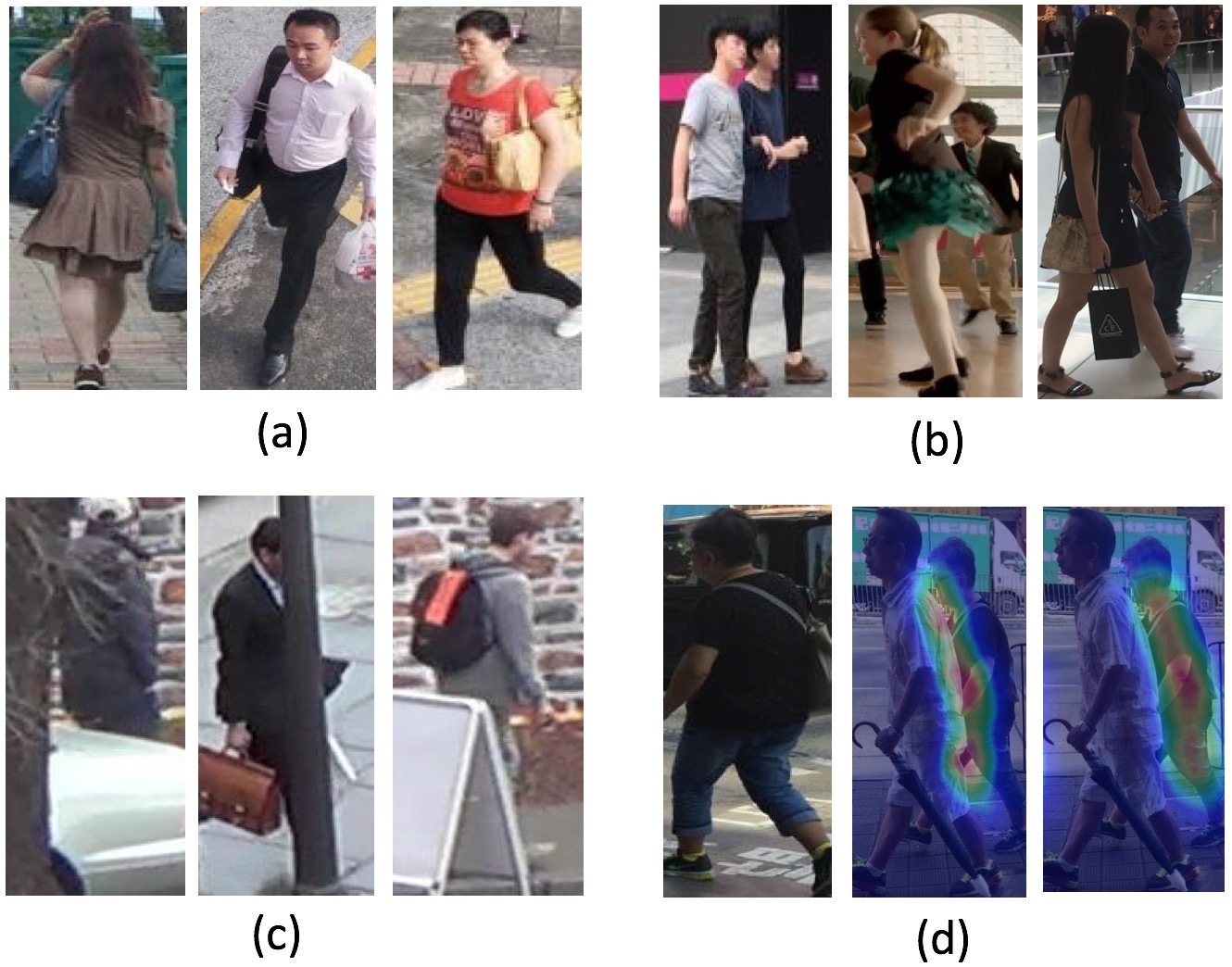}
  \caption{Typical samples in the (a) traditional Re-ID, (b) Pedestrian-Interference Re-ID, and (c) Occluded Re-ID from Market-1501~\cite{zheng2015scalable}, our constructed PI-CUHK-SYSU, and Occluded-DukeMTMC~\cite{ristani2016MTMC,zheng2017unlabeled}, respectively. 
  (d) shows the query image (the first sample) and, the comparison of visualization results between the Occluded Re-ID method Foreground-aware Pyramid Reconstruction~\cite{He_2019_ICCV} (the second one) and our PISNet (the third one)
  }
 \label{fig:hard_case}
\end{center}
\end{figure*}

Re-IDentification (Re-ID) aims to identify the same person across a set of images from nonoverlapping camera views, facilitating cross-camera tracking techniques used in video surveillance for public security and safety. 
In general, person Re-ID is considered to be the next high-level task after a pedestrian detection system.
Therefore, as shown in Figure~\ref{fig:hard_case}(a), the basic assumption of Re-ID is that the detection model can provide a precise and highly-aligned bounding box for each individual. 
However, in a crowded scene, off-shelf-detectors may draw a bounding box containing multiple people, as shown in Figure~\ref{fig:hard_case}(b).
This means the cropped images contain both the target and the interference pedestrians.
The interference pedestrian makes the feature ambiguous to identify the target person, which might lead to wrong retrieval results.
We call this the Pedestrian-Interference person Re-IDentification (PI Re-ID) problem.

We observe that mutual occlusion of pedestrians often occurs in PI Re-ID. 
Recent works~\cite{He_2019_ICCV,Miao_2019_ICCV,vpm_cvpr2019,Fan_2019} have well studied the Occluded Re-ID problem.
However, in their setting of Occluded Re-ID, the person images are mainly occluded by obstructions like cars, trees, or shelves. 
This is also reflected in the existing benchmarking Occluded Re-ID datasets, most of which consist of non-pedestrian occlusion, as shown in Figure~\ref{fig:hard_case}(c). 
The performance of their approaches degrades if directly applied to PI Re-ID, as shown in the second sample of Figure~\ref{fig:hard_case}(d).
Because they only focus on reducing the influence caused by obstructions and do not specifically consider the interference between pedestrians in a cropped image.
Moreover, they  do  not explicitly  learn  to  draw  a  precise  boundary  between  two overlapping people so that the extract features are corrupted by each other.
As for PI Re-ID, it is different from Occluded Re-ID in two aspects: 1) PI Re-ID focuses on the pedestrian interference, which is more confusing than the non-pedestrian obstructions.
2) PI Re-ID aims to re-identify all the pedestrians appearing in a cropped image, which might be interfered with the background pedestrians or the pedestrian occlusion.
Therefore, our setting is more challenging than Occluded Re-ID. 
Moreover, our setting is more practical in the crowded situation (\textit{e.g.}, airports, railway stations, malls, and hospitals), where people always share overlapping regions under cameras. 
 
\begin{figure*}
\begin{center}
\includegraphics[width=10cm]{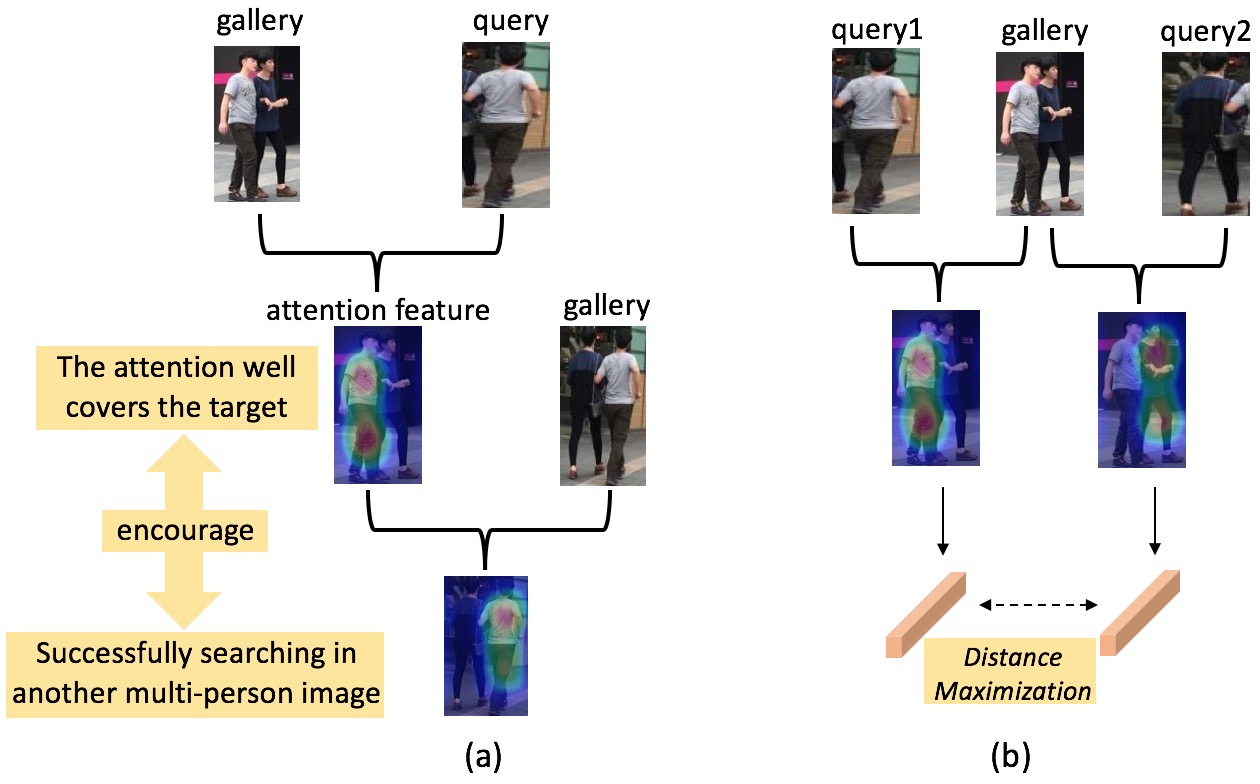}
   \caption{(a) Enhance the location accuracy of the attention by GRAM. (b) Facilitate the feature distinctiveness of the attention by MPSL}
 \label{fig:simple_illus}
\end{center}

\end{figure*}

To retrieve a person in the PI Re-ID setting, the extracted features should ensure 1) \textbf{location accuracy}: the strong activation on all the regions of targets, 2) \textbf{feature distinctiveness}: the trivial feature corruption by other pedestrians. 
To achieve this goal, we propose a novel deep network termed Pedestrian-Interference Suppression Network (PISNet), which consists of a backbone Fully Convolutional Network (FCN), a Query-Guided Attention Block (QGAB) and a Guidance Reversed Attention Module (GRAM).
First, FCN is utilized to extract features for person images.   
Since the target feature, in a gallery image containing multi-person information, differs on the query,
QGAB is designed to enhance the feature of the target in the gallery and suppress that of interference pedestrians, under the guidance of the query. 
On the one hand, as shown Figure~\ref{fig:simple_illus}(a), for encouraging the \textbf{location accuracy} of the attention, our motivation is that, if the attention well covers the regions of the target, the attention feature can be further utilized as the guidance to search the target in other multi-person images. 
Therefore, GRAM leverages the refined gallery features to guide other multi-person features to formulate attention for targets.
On the other hand, as shown in Figure~\ref{fig:simple_illus}(b), to facilitate the \textbf{feature distinctiveness} of the attention learning, PISNet utilizes the Multi-Person Separation Loss (MPSL) to maximize the distance between the features, which are extracted from the same gallery but guided by different queries.
In addition, as shown in the third sample of Figure~\ref{fig:hard_case}(d), our PISNet is more capable of depressing the pedestrian interference than the Occluded Re-ID method.

Our {\bf contributions} are listed as follows: 1) To the best of our knowledge, it is the first work that particularly addresses the problem of PI Re-ID. 
2) We propose a Pedestrian-Interference Suppression Network (PISNet), which utilizes a query-guided approach to extract the target feature. The involving GRAM and MPSL further promote the location accuracy and the feature distinctiveness of the attention learning, respectively.
3) Since the existing benchmarks largely ignored this problem, we contribute two new datasets, which are specifically designed for this problem with a great deal more pedestrian-interference instances.
Our experimental results on these two datasets show that the proposed model is effective in addressing the PI Re-ID problem, yielding significant improvement over representative Re-ID methods applied to the same problem.
The experimental results also demonstrate the generalization ability of our method on the general Re-ID datasets (Market-1501 and DukeMTMC-ReID).

\section{Related Work}

\subsection{Person Re-ID} 
Generally, Person Re-ID can be divided into two steps: calculating a feature embedding and performing feature matching under some distance metric~\cite{weinberger2006distance,li2013learning,khamis2014joint}. We mainly review the former including both handcrafted feature~\cite{khamis2014joint,koestinger2012large,li2013locally,ma2012bicov} and learned feature~\cite{li2014deepreid,zhao2014learning,cheng2016person,hermans2017defense,liao2017triplet,zheng2018pyramidal} approaches. 

In recent years, Re-ID has witnessed great progress owing to the prevailing success of convolutional neural networks (CNNs) in computer vision. 
However, simply applying CNNs to feature extraction may not yield ideal Re-ID performance due to many problem-specific challenges such as partial body, background perturbance, view point variation, as well as occlusion/misalignment. 
Combining the image-level information with the human-part information can enhance the robustness of Re-ID models.
Moreover, many part-based approaches have achieved considerable improvement~\cite{gray2008viewpoint,prosser2010person,liao2015person,ma2013domain,zheng2013reidentification,cheng2016person,su2017pose,wei2017glad,zheng2017pose,zhu2017part,suh2018part,zhao2017deeply,zhao2017spindle}. We refer readers to \cite{zheng2016person} for a more comprehensive review.

\subsection{Attention Mechanisms in Person Re-ID}
Several studies leverage attention mechanisms to address the misalignment problem in person Re-ID. 
For example, Chen et al.~\cite{Chen_2019_ICCV} propose an attentive but diverse network which consists of a pair of complementary attention modules, focusing on channel aggregation and position awareness, respectively.
%
%
Si et al.~\cite{si2018dual} use an inter-class and an intra-class attention module to capture the context information for person Re-ID in video sequences.
Li et al.~\cite{Li_2018_CVPR} leverage hard region-level and soft pixel-level attention, which can jointly produce more discriminative feature representations.
Xu et al.~\cite{xu2018attention} utilize pose information to learn attention masks and then combine the global with the part features as feature embeddings.

Previous methods~\cite{Chen_2019_ICCV,Xia_2019_ICCV,xu2018attention} leverage attention mechanisms to enhance the feature of human bodies.
In contrast, in our proposed setting, images contain other pedestrians, which severely corrupt the feature of a target.  
Since they cannot distinguish between the target and interference pedestrians, directly applying their approaches will cause the severe corruption of the target feature.
. 

\subsection{Occluded Re-ID} 
Some related works for the Occluded Re-ID have been well studied.
Zheng et al.~\cite{zheng_partial} propose an Ambiguity sensitive Matching Classifier (AMC) and a Sliding Window Matching (SWM) model for the local patch-level matching and the part-level matching, respectively.
He et al.~\cite{he2018deep} propose a Deep Spatial Feature Reconstruction (DSR) model for the alignment-free matching, which can sparsely reconstruct the spatial probe maps from spatial maps of gallery images. 
He et al.~\cite{He_2019_ICCV} further present a Spatial Feature Reconstruction (SFR) method to match different sized feature maps for the Partial Re-ID.
Miao et al.~\cite{Miao_2019_ICCV} propose the Pose-Guided Feature Alignment (PGFA), which introduces the pose estimation algorithm to enhance the human part feature in an occlusion image.
Sun et al.~\cite{vpm_cvpr2019} propose a self-supervision model called Visibility-aware Part Model (VPM), which can perceive the visibility of regions.
Fan et al.~\cite{Fan_2019} propose a spatial-channel parallelism network (SCPNet), which enhances the feature of a given spatial part of the body in each channel of the feature map.
%

These methods ignore the interference between pedestrians within a cropped image. 
Therefore, they cannot well address the PI Re-ID problem, where the large proportion of the pedestrian interference exists.
In contrast, in this paper, we focus on suppressing the pedestrian interference, by learning the query-guided attention with the location accuracy and the feature distinctiveness.

\section{Pedestrian-Interference Suppression Network}
\label{sec:method}

In this work, we assume that in PI Re-ID a query image contains only a single person and the task is to match this query with a gallery consisting of the pedestrian interference. 
This is based on a practical scenario where a human operator has manually cropped the human body and sent a query to a Re-ID system to search for the same person in another camera view. 
In this section, we first give an overview of our framework, and then describe more details for each component individually. 

\subsection{Overview} 
\begin{figure*}[!tb]
\begin{center}
\includegraphics[width=12cm]{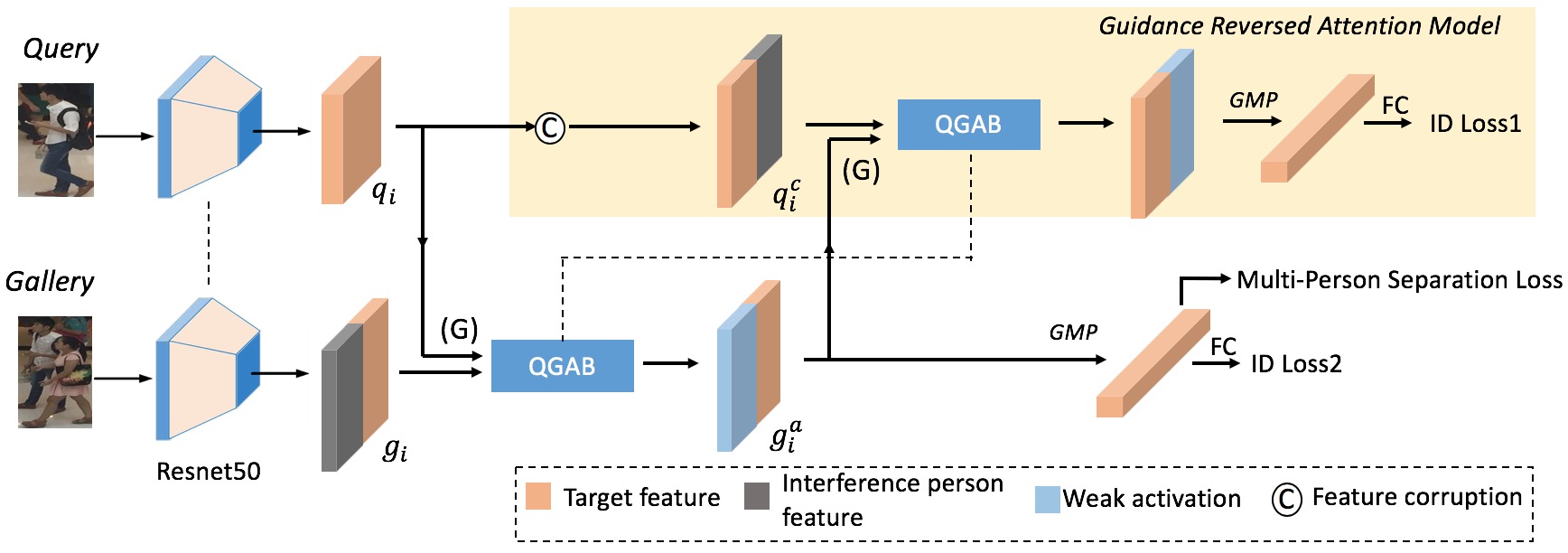}
   \caption{Illustration of our Pedestrian-Interference Suppression Network (PISNet).
   For further clarity, the target feature represents the same ID information to the query. 
   PISNet consists of (1) a backbone Fully Convolutional Network (FCN),
    (2)	a Query-Guided Attention Block (QGAB), and
    (3)	a Guidance Reversed Attention Module (GRAM).
    FCN can extract features for the query and the gallery.
    QGAB leverages the query feature as the guiding feature (single-person) to formulate attention on the gallery feature (with pedestrian interference). 
    GRAM plays a role in encouraging QGAB to enhance the feature on the regions of a target. 
    The Multi-person Separation Loss promotes the attention to draw a more precise boundary for overlapping instances. 
    $g_{i}$ and $q_{i}$ denote the feature map of a gallery and a query, respectively.
    $g^{a}_{i}$ and $q^{c}_{i}$ are the refined gallery feature and the corrupted query feature, respectively. 
    GMP denotes the Global Max Pooling. 
    QGABs share the same parameters.
    (G) denotes the feature as the guidance to QGAB. 
    GRAM is only used in the training stage
   }
 \label{fig:sranet}
\end{center}
\end{figure*}

As shown in Figure~\ref{fig:sranet}, PISNet consists of  (1) a backbone Fully Convolutional Network (FCN), 
(2)	a Query-Guided Attention Block (QGAB), and
(3)	a Guidance Reversed Attention Module (GRAM).
For each forward propagation in the training stage, we pair a gallery image with a query image. 
FCN can extract features for the query and the gallery.
QGAB finds the common regions between the query and gallery feature maps, and then enhances the common feature in the gallery feature. 
For encouraging the location accuracy of the attention, GRAM aims to guarantee that the refined gallery feature has strong attention on all the regions of the target.
For the feature distinctiveness of the attention, the Multi-person Separation Loss (MPSL) magnifies distance of the features from the same gallery but guided by different queries.
In addition, GRAM is ignored in the testing stage.

\subsection{Query-Guided Attention Block} 

\begin{figure*}[!tb]
\begin{center}
\includegraphics[width=8cm]{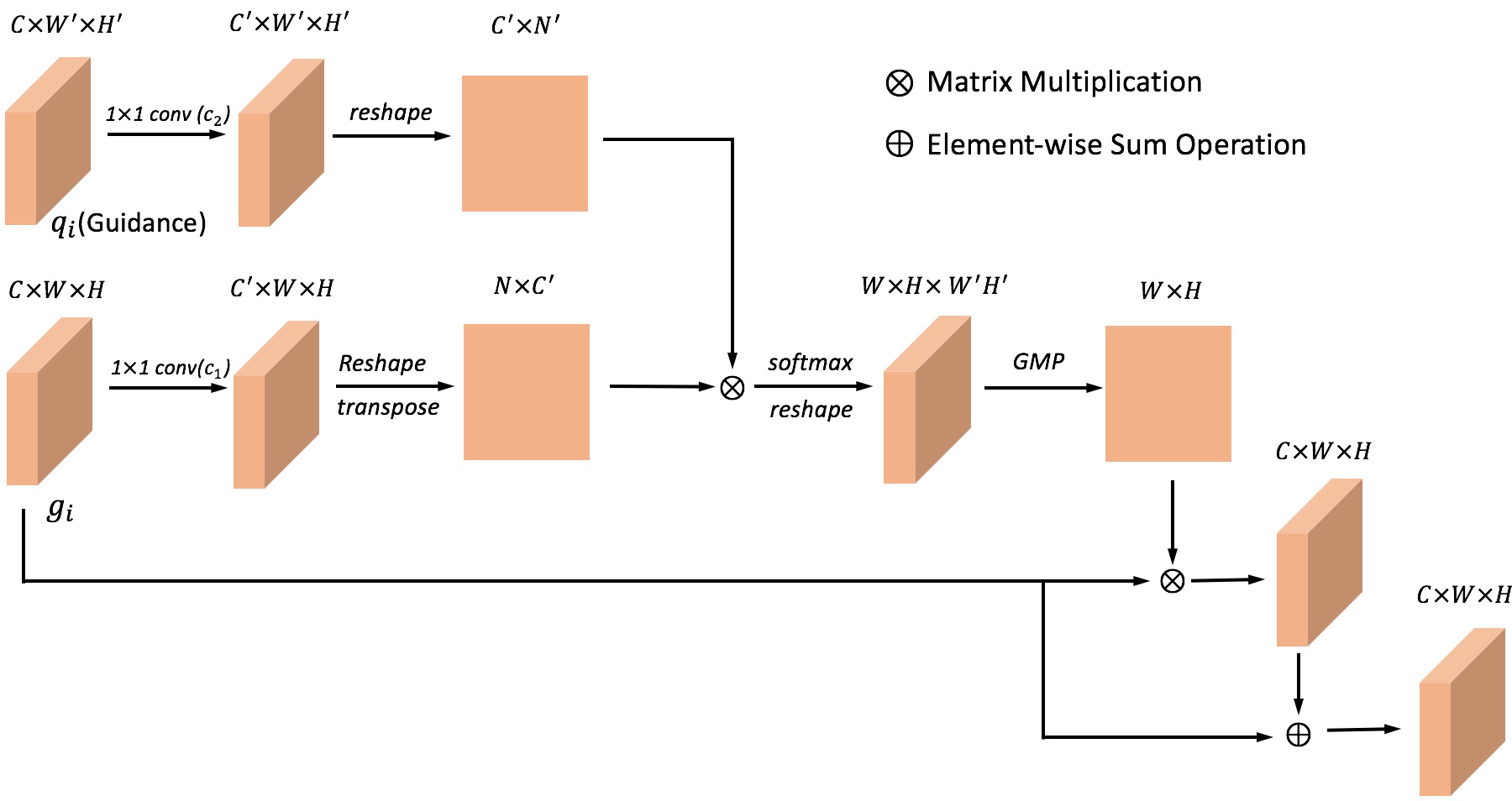}
   \caption{Illustration of our proposed Query-Guided Attention Block (QGAB).
$g_{i}$ denotes the feature map of a gallery.
$q_{i}$ is the feature map of a query.
The feature maps are shown as the shape of their tensors.
$W$ and $H$ are the width and height of the gallery feature map. 
$W^{'}$ and $H^{'}$ are the width and height of the query feature map.
$C$ is the number of channels after the backbone. 
GMP denotes the Global Max Pooling
%
}
 \label{fig:qgab}
\end{center}
\end{figure*}

QGAB is depicted in Figure~\ref{fig:qgab}. 
The main goal of this block is to search for spatial similarity between the query and the multi-person gallery.
The inputs of QGAB are the query and gallery feature maps. 
The query is used as the guidance.
The output is the spatially enhanced gallery feature. 
The spatial similarity calculates the inner product of the features from gallery and query branch first, after which Global Max Pooling (GMP) in the channel dimension is applied to formulate a pixel-wise attention matrix. 
This matrix then is multiplied with the gallery feature in order to enforce a spatial similarity search between the query and gallery feature maps. 
The overall process of this feature enhancement is formulated as:

\begin{equation}
QGAB(g_{i},q_{i}) = GMP\big(Softmax({c_{1}(g_{i})}^{T} \times {c_{2}(q_{i})})\big) \times g_{i} + g_{i},
\end{equation}

\noindent where $g_{i}$ is the multi-person feature (gallery), $q_{i}$ is the single-person feature (query), $c_{1}$ and $c_{2}$ are convolutional layers, $GMP$ is the Global Max Pooling in the channel dimension and $\times$ denotes matrix multiplication.

\begin{figure*}
\begin{center}
\includegraphics[width=8cm]{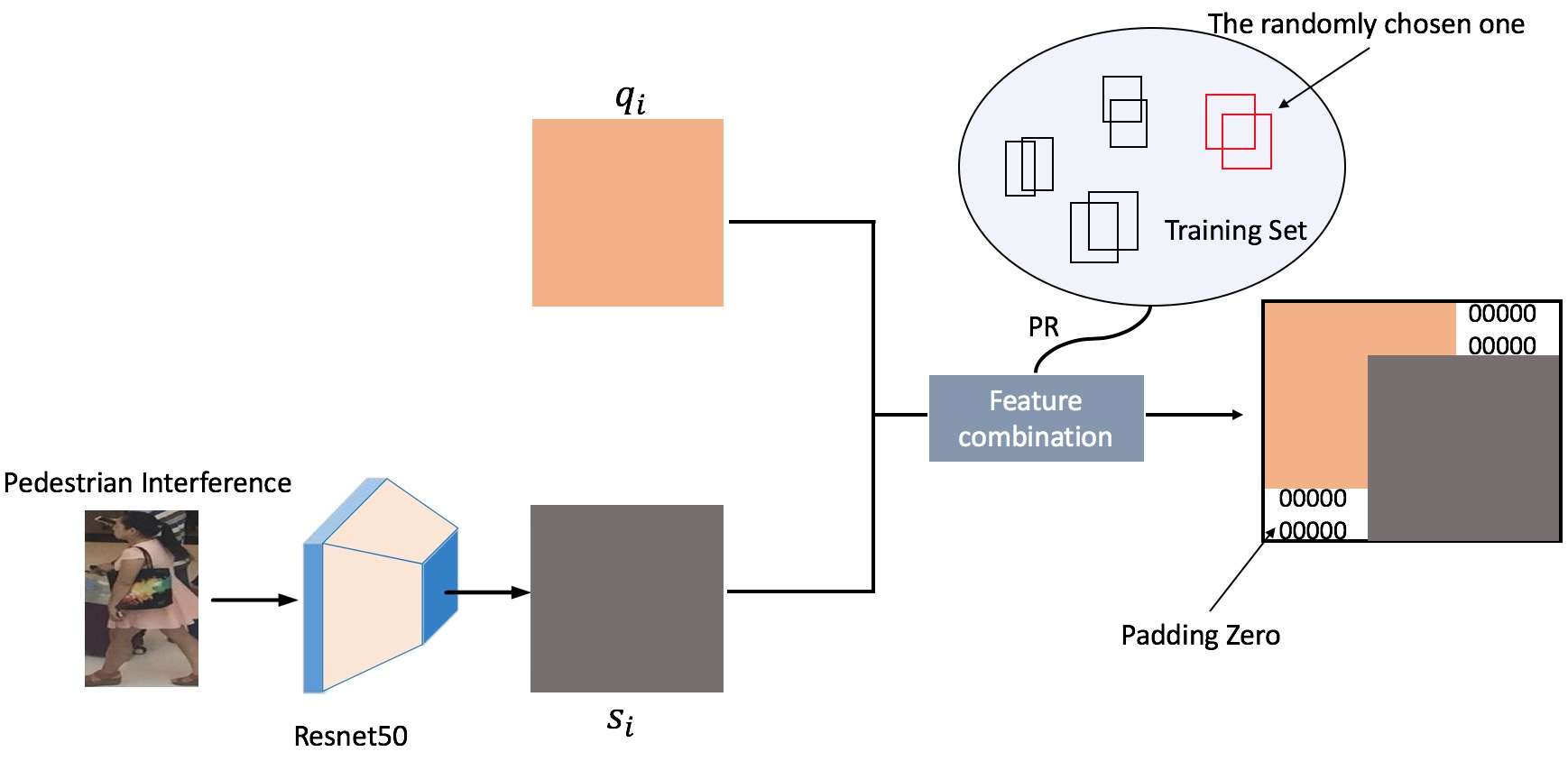}
   \caption{Illustration of the feature corruption.
$q_{i}$ denotes the query feature map.
$s_{i}$ denotes the feature map of the sampled single-person image.
$PR$ denotes the function that extracts the relative position relationship of ground-truth boxes from a multi-person image.
Two features are combined following the relative position relationship
}
 \label{fig:fc4}
\end{center}

\end{figure*}

\subsection{Guidance Reversed Attention Module} 

GRAM aims to guarantee that the refined gallery feature has the strong attention on all the regions of the target. 
As shown in Figure~\ref{fig:simple_illus}(a), our motivation is that a well-refined gallery feature can be used as the guidance to formulate the attention for another gallery feature containing the pedestrian interference. 
For example, if a gallery contains IDs of A and B. 
Using the query image of A, the well-refined gallery feature will have the strong activation on regions of A while the feature of B is suppressed. 
Therefore, the refined feature should be capable of serving as the guidance to formulate attention for another gallery containing person A. 
In this new attention mask, we expect the feature of A is still enhanced and the feature of another pedestrian is suppressed. 
The attention formulation for person A in these two feature maps is encouraged by each other.

Therefore, as shown in Figure~\ref{fig:sranet}, we utilize the feature, which is refined by our query-guided attention operation, as the guidance feature to formulate the spatial attention on another gallery feature map.  
In order to reduce the labour of the data collection, we construct the new gallery by a feature corruption operation.
Specifically, we randomly select a single-person image and a gallery image with the pedestrian interference.
We can extract the single-person feature from the former and the relative position relationship of the involving ground-truth bounding boxes from the latter.
Then we corrupt the query feature by combining it with the single-person feature.
Specifically, as shown in Figure~\ref{fig:fc4}, following the relative position relationship, we put two feature maps on the corresponding positions to generate a multi-person feature map and we pad the remaining regions with zero.
The process of this feature corruption is formulated as:

\begin{equation}
FC(q_{i}, s_{i}, m) = Combine(q_{i}, s_{i}, PR(m)),
\end{equation}

\noindent where $s_{i}$ is the feature map of the sampled single person image, $m$ denotes the image with pedestrian interference, $PR$ denotes the function that extracts the relative position relationship of ground-truth boxes from $m$, and $Combine$ denotes the function that can combine features depending on the relative position relationship of bounding boxes.

Then we input the corrupted query feature and the refined gallery feature into QGAB. 
In contrast to the last QGAB operation, we reverse the roles of two features. 
The refined gallery feature is served as a query feature that can guide QGAB to enhance the target feature in the corrupted query features. 
The overall process of reversed feature enhancement is formulated as:

\begin{equation}
QGAB^{r}(g^{a}_{i},q^{c}_{i}) = GMP\big({Softmax(c_{1}(q^{c}_{i})}^{T} \times {c_{2}(g^{a}_{i})})\big) \times q^{c}_{i} + q^{c}_{i},
\end{equation}

\noindent where $g^{a}_{i}$ is the refined gallery feature, $q^{c}_{i}$ is the corrupted query feature, and $c_{1}$ and $c_{2}$ share parameters with the last QGAB.

\begin{figure*}[!tb]
\begin{center}
\includegraphics[width=10cm]{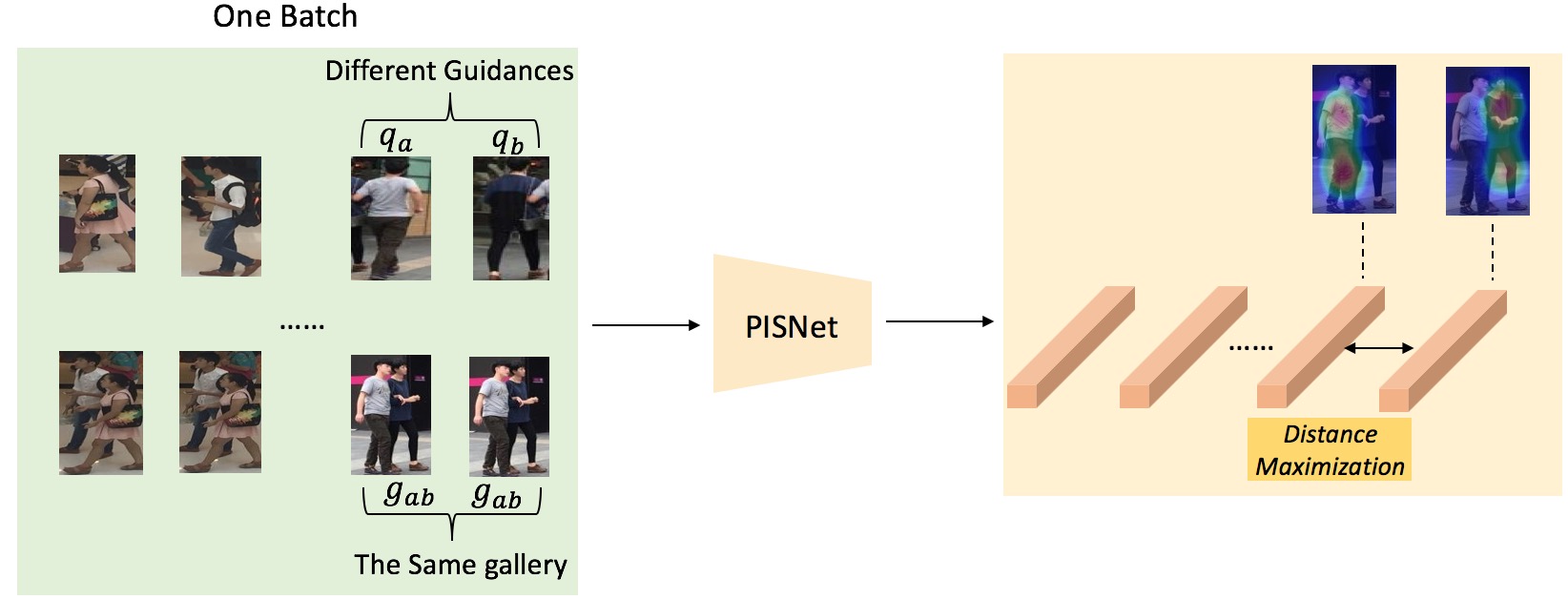}
   \caption{The computation process of the Multi-Person Separation Loss.
   $q_{a}$ and $q_{b}$ denote the queries.
   $g_{ab}$ is the gallery.
   The subscripts represent the IDs that appear in the image.
   In one batch, we pair a multi-person gallery image with different query images as the guidances. 
   To promote the feature distinctiveness of the attention, the distance between the refined features guided by different query images should be maximized
   }
 \label{fig:mpsloss}
\end{center}

\end{figure*}

\subsection{Multi-Person Separation Loss} 

In a pedestrian-interference image, people always share an overlapping area of their body.
This is the key reason that causes the failure detection. 
Moreover, it also improves the difficulty of the attention learning. 
Therefore, we conduct the feature distinctiveness enhancement by the Multi-Person Separation Loss for further guaranteeing the purity of refined features.

As shown in Figure~\ref{fig:simple_illus}(b), we expect that the refined feature should have a large distance to the feature guided by another query image with a different person ID. 
For example, if a gallery image contains A and B, given the query image of A, we expect to extract the pure feature of A while suppressing the feature of B. 
In contrast, given the query image of person B, the pure feature of B should be extracted. 
In order to achieve this goal, as shown in Figure~\ref{fig:mpsloss}, we first construct the image batch for training, where a multi-person gallery image is paired with different query images as the guidances.
Then, the distances can be minimized and maximized, respectively, by the Multi-Person Separation Loss, which is given by, 

\begin{equation}
 \begin{aligned}
L_{m} = max(0, c + dist(QGAB(g_{ab},q_{a}), QGAB(g_{ab},q_{b})) - \\
dist(QGAB(g_{ab},q_{a}), q_{a})),
 \end{aligned}
\end{equation}

\noindent where $dist$ is the cosine distance and $c$ denotes the margin coefficient.
We should maximize the distance between $QGAB(g_{ab},q_{a})$ and  $QGAB(g_{ab},q_{b})$ and meanwhile minimize the distance between $QGAB(g_{ab},q_{a})$ and $q_{a}$, where the subscripts represent the IDs that appear in the image.

\subsection{Overall  Objective  Function}

We utilize the cross entropy loss for both the gallery branch and GRAM, which is denoted as $L_g$ and $L_q$, which is corresponding to $ID\ Loss_{1}$ and $ID\ Loss_{2}$, respectively, in Figure~\ref{fig:sranet}.

\begin{equation}
L_{g} = CE(\hat{y},y),
\end{equation}

\begin{equation}
L_{q} = CE(\bar{y},y),
\end{equation}

\noindent where $CE$ denotes the cross-entropy loss, $\hat{y}$ and $\bar{y}$ denote the prediction ID in the gallery branch and GRAM, respectively, and $y$ is the ground-truth ID.
By combining with the Multi-Person Separation Loss, the final loss for the network is formulated as

\begin{equation}
L_{final} = L_{g} + \alpha*L_{q} + \beta*L_{m},
\end{equation}

\noindent where $\alpha$ and $\beta$ are the coefficients to balance the contributions from the latter two losses. 

\subsection{Implementation Details}
To implement our proposed model, we adopt Resnet-50~\cite{he2016deep} as our basic CNN for feature extraction, which is pretrained on ImageNet.
We first train the backbone on the single-person images using all training tricks in the strong baseline~\cite{luo2019bag}. Then we add QGAB on the top of the Siamese Network. 
Both $c_{1}$ and $c_{2}$ are $1 \times 1$ convolutional layers with 1024 channels. 
Then we freeze the backbone network and train QGAB by pairing the multi-person images with single person ones. 
The batch size of samples for training is 64. 
The SGD optimizer is applied, with a learning rate of 0.00035. 
They are decayed by 0.1 after 20 epochs, and the training stops at 60 epochs. 
Parameters for the final loss function are $\alpha=1.0$ and $\beta=0.5$.

\section{Experiments}

\subsection{Datasets and Evaluation Metrics}

To demonstrate the effectiveness of our model on the Person-Interference Re-ID problem, we carry out the experiment on our constructed PI-PRW and PI-CUHK-SYSU dataset. 
Besides, in order to prove the generalization ability of our method on single-person images, we also evaluate the proposed PISNet on the another two datasets: Market-1501~\cite{zheng2015scalable} and DukeMTMC-ReID~\cite{ristani2016MTMC,zheng2017unlabeled}.

\noindent \textbf{PI-PRW} is derived from the PRW~\cite{Zheng_2017_CVPR} dataset.
We use the off-the-shelf detector Faster R-CNN~\cite{ren2015faster} to perform pedestrian detection. Then we select the bounding boxes with multiple pedestrians.
The selection criterion is: 1) At least 70\% area of each ground-truth bounding box should be contained in the multi-person boxes. 
2) The contained part of bounding boxes is at least 0.3 times the size of multi-person boxes in order to ensure the degree of the person interference. 
3) Each bounding box has the overlapping area with any other ones.  
We get 1792 multi-person images with 273 IDs for training and 1258 multi-person gallery images and 211 single person query images for testing. 
Besides, in order to get closer to the actual scene, we add another 10000 single-person images in the test set as gallery images.  

\noindent \textbf{PI-CUHK-SYSU} is derived from the CUHK-SYSU~\cite{Xiao_2017_CVPR} dataset. 
We get multi-person cropped images following the same procedure in PI-PRW, resulting 3600 multi-person images for training with 1485 IDs and 3018 multi-person gallery images and 1479 single person query images for testing.
We also add another 10000 single-person images in the test set as gallery images.
More details of PI-PRW and PI-CUHK-SYSU can be referred to our supplementary material.

\noindent \textbf{Evaluation Metrics.} We use Cumulative Matching Characteristic (CMC) curves and mean average precision (mAP) to evaluate the quality of different Re-ID models. All the experiments are performed in a single query setting.

\begin{table}
\begin{center}
\caption{\label{tbl:pre}Comparison results ($\%$) on PI-PRW and PI-CUHK-SYSU dataset at $4$ evaluation metrics: \textit{rank 1}, \textit{rank 5}, \textit{rank 10}, \textit{mAP} where the bold font denotes the best method. 
The methods in the 1st group are proposed for the traditional Re-ID problem.
The methods in the 2nd group are proposed for the multi-label learning. 
The 3rd group is the methods of Occluded Re-ID.
The 4th group is our method}

{\footnotesize
\begin{tabular}{l|ccc|c|ccc|c}
\hline
\hline
\multirow{2}{*}{Method}	&	\multicolumn{4}{c|}{PI-PRW}	&			\multicolumn{4}{c}{PI-CUHK-SYSU}			\\
\cline{2-9}
	&	\textit{rank1} 	&	\textit{rank 5}	&	\textit{rank 10}	&	\textit{mAP}	&\textit{rank1} 	&	\textit{rank 5}	&	\textit{rank 10}	&	\textit{mAP}\\
\hline\hline									
HA-CNN~\cite{Li_2018_CVPR}           &	32.4 & 56.9  & 68.0 &  32.0	 &	71.3 	& 82.0	 	& 87.5 & 65.3\\
PCB~\cite{sun2018beyond}	         &  31.3 & 55.1 &  67.5	&  30.2	 &	70.1 	& 80.4 & 86.9  & 63.1  \\
Strong Baseline~\cite{luo2019bag}   &  34.7	& 59.4	& 70.3	&  36.0  & 72.5 & 83.9  &  88.2 & 70.1 \\
PyramidNet~\cite{Zheng_2019_CVPR}	 &	35.9	& 60.2	& 70.1	&	37.0 &	73.1 & 83.5	 	& 87.9 & 70.5  \\
ABD-Net~\cite{Chen_2019_ICCV}	         &	35.4	& 59.9	& 69.7	&  36.3 &	72.9 &	82.6 	& 87.5	& 70.4 \\
QAConv~\cite{Shengcai_QG}  &  36.0   &  61.2    & 	70.9    &  38.2  &  73.2 & 84.7  & 88.3 &  70.9 \\
\hline	
HCP~\cite{hcp_2016}   &   30.2 & 49.7 & 61.2	& 29.6 & 67.2 &	75.3 	& 83.5		& 61.9 \\
LIMOC~\cite{Yang_2016_CVPR}    &  32.9 & 52.4 & 63.3	& 32.6 & 69.1 & 78.2	& 85.3		& 65.2 \\
\hline
FPR~\cite{He_2019_ICCV}   &  36.3	 &	60.7 	&	70.4	&  37.9	 &	73.7 	&	85.0 	&	89.1	& 71.2 \\
AFPB~\cite{occluded_icme} &  34.1	 &	58.2 	&	67.2	&  35.1	 &	70.7 	&	83.2 	&	87.3	& 68.3 \\
\hline
Ours	                & \textbf{42.7} & \textbf{67.4} & \textbf{76.2}	&  \textbf{43.2}  & \textbf{79.1} &  \textbf{88.4}	 &	\textbf{91.9}	& \textbf{76.5} \\
\hline
\end{tabular}}

\end{center}

\end{table}

\subsection{Results Comparison}

\noindent \textbf{Results on PI-PRW and PI-CUHK-SYSU.} 
We first compare the proposed approach with the existing methods on the two proposed PI Re-ID datasets. 
Table~\ref{tbl:pre} shows the result of our method and previous works. 
The compared methods (including six existing representative Re-ID models, two multi-label learning approaches, and two Occluded Re-ID methods) are listed in the table. 
These results show: (1) Among existing methods, the Occluded Re-ID model FPR is superior.
For example, FPR achieves 36.3\% Rank-1 accuracy and 37.9\% mAP on PI-PRW, which outperforms all the previous Re-ID methods.
This is because, similar to our method, FPR~\cite{He_2019_ICCV} leverage query feature maps as multi-kernels to calculate the spatial affinity with the gallery feature maps, and then enhance the common features in the gallery features. 
(2) The performance of LIMOC~\cite{Yang_2016_CVPR} and HCP~\cite{hcp_2016} proposed for the multi-label learning is ordinary.
For example, compared to the strong baseline~\cite{luo2019bag}, HCP~\cite{hcp_2016} is less by -4.5\% Rank-1 accuracy and -3.5\% mAP on PI-PRW.
(3) Our new model PISNet outperforms all competitors by significant margins.
For example, PISNet achieves 42.7\% Rank-1 accuracy and 43.2\% mAP on PI-PRW and 79.1\% Rank-1 accuracy and 76.5\% mAP on PI-CUHK-SYSU.
This is because our proposed method explicitly utilizes query information and learn a more precise boundary between pedestrians by GRAM and MPSL.

\begin{table}
\begin{center}
\caption{\label{tbl:component}Component analysis of the proposed method on the PI-PRW and PI-CUHK-SYSU datasets ($\%$)
}
{
\scriptsize
\begin{tabular}{l|ccc|c|ccc|c}

\hline
\hline

\multirow{2}{*}{Method}	&	\multicolumn{4}{c|}{PI-PRW}	&			\multicolumn{4}{c}{PI-CUHK-SYSU}			\\
\cline{2-9}

	&	\textit{rank1} 	&	\textit{rank 5}	&	\textit{rank 10}	&	\textit{mAP}	&\textit{rank1} 	&	\textit{rank 5}	&	\textit{rank 10}	&	\textit{mAP}\\
\hline\hline									
Baseline                &  34.7	& 59.4	& 70.3	&  36.0  & 72.5 & 83.9  &  88.2 & 70.1 \\
Baseline + QGAB	        &  38.9	& 61.5	& 72.4	&  38.0  & 73.9 & 85.0  &		89.1&  72.3 \\
Baseline + QGAB + MPSL	&  39.7	& 63.2	& 74.1	&  40.1  & 76.2	& 87.1	&		91.4&  74.2 \\
Baseline + QGAB + GRAM	    &  41.8	& 66.1	& 75.2	&  42.4  &  77.9 & 87.5	 	& 91.0 & 75.0 \\
Baseline + QGAB + GRAM + MPSL &  42.7	& 67.4	& 76.2	&  43.2  & 79.1 &  88.4	 &	91.9	& 76.5 \\
\hline
\end{tabular}
}

\end{center}


\end{table}

\renewcommand{\multirowsetup}{\centering}
\begin{table}[tb]\footnotesize

  \begin{center}
  
  \caption{\label{tab:single}Comparison results ($\%$) on the Market-1501 and DukeMTMC-ReID datasets.
  $N_{f}$ is the number of features used in the inference stage. 
  The methods in the 1st group are proposed for the traditional Re-ID problem.
  The 2nd group is the state-of-the-art methods of Occluded Re-ID.
  The 3rd group is our method
  }
  
  \begin{tabular}{ cc|cc|cc}
\hline
   &       & \multicolumn{2}{c|}{Market1501} & \multicolumn{2}{c}{DukeMTMC-ReID}	 \\
   Method & $N_f$	   & r = 1 	& mAP	&r = 1 	& mAP 	 \\
 	\hline
	\hline
    PIE~\cite{zheng2017pose}          & 3  &87.7	&69.0	&79.8	&62.0 \\
    SPReID~\cite{kalayeh2018human}    & 5  & 92.5 & 81.3	& 84.4	&71.0 \\
    MaskReID~\cite{qi2018maskreid}    & 3  &90.0	&75.3	&78.8	&61.9 \\
    MGN~\cite{wang2018learning}       &1& 95.7 & 86.9& 88.7 & 78.4 \\
    SCPNet~\cite{Fan_2019}       & 1  & 91.2	&75.2	&80.3	&62.6		\\
    PCB~\cite{sun2018beyond}          & 6  & 93.8	&81.6	&83.3	&69.2	\\
    Pyramid~\cite{Zheng_2019_CVPR}     & 1 & 92.8 &82.1	&-	&-		\\
    Pyramid~\cite{Zheng_2019_CVPR}     & 21 & 95.7 &88.2	&89.0	&79.0	\\
    HA-CNN~\cite{Li_2018_CVPR}    & 4 & 91.2	& 75.7	&80.5	&63.8	\\
    ABD-Net~\cite{Chen_2019_ICCV} & 1 & 95.6	& 88.3	&89.0	&78.6\\
    Camstyle~\cite{zhong2019camstyle} &  1 &88.1 	&68.7	&75.3	&53.5	\\
    PN-GAN~\cite{Qian_2018_ECCV}        &  9 &89.4 	&72.6	&73.6	&53.2	\\
    IDE~\cite{zheng2018discriminatively}  & 1  & 79.5	& 59.9	& -	&-		\\
    SVDNet~\cite{Sun_2017_ICCV}  & 1  & 82.3	& 62.1	& 76.7	&56.8		\\
    TriNet~\cite{hermans2017defense}  & 1  & 84.9	& 69.1	& -	& -		\\
    SONA~\cite{Xia_2019_ICCV} & 1  & 95.6	& 88.8	& 89.6	& 78.2	\\
    \hline
    FPR~\cite{He_2019_ICCV}   &1& 95.4& 86.5& 88.6&78.2\\
    PGFA~\cite{Miao_2019_ICCV}	&1& 91.2& 76.8& 82.6&65.5\\
    \hline
    Baseline &1   & 94.5 &85.9& 86.4&76.4        \\
    \textbf{Ours}    &  1 & 95.6& 87.1 & 88.8&   78.7 \\
\hline

  \end{tabular}
  \end{center}

\end{table}

\begin{figure*}
\begin{center}

\includegraphics[width=12cm]{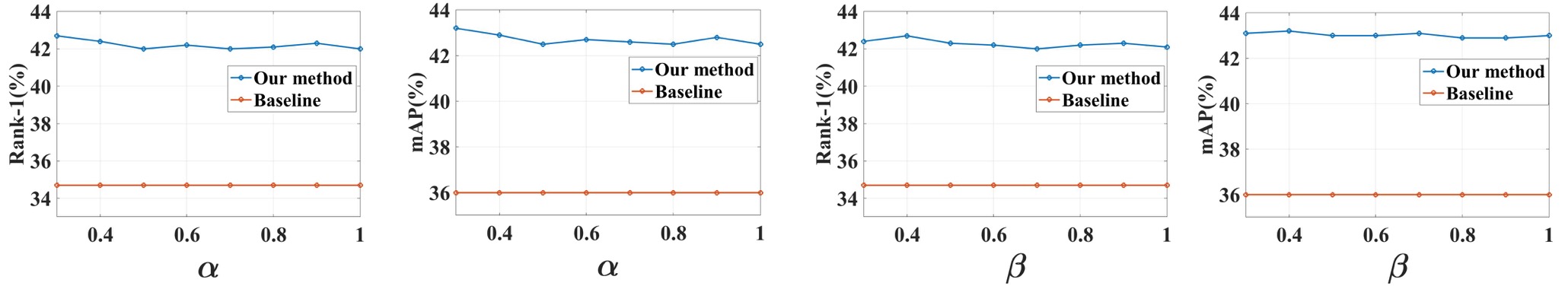}
   \caption{Evaluation of different parameters of PISNet using Rank-1 and mAP accuracy on the PI-PRW dataset ($\%$)
   }
 \label{fig:sa}

\end{center}
\end{figure*}

\subsection{Further Analysis}

\noindent \textbf{Contributions of Individual Components.} In Table~\ref{tbl:component}, we evaluate the three components on how they contribute to the full model. 
The results show that all of them are effective on their own (each outperforms all the compared methods). 
Moreover, when combined, the best performance is achieved. 
This validates our design consideration in that they are complementary and should be combined.

\noindent \textbf{Does PISNet Perform well on the General Re-ID Dataset?} We also apply our method on the general Re-ID datasets, Market-1501 and DukeMTMC-ReID. 
The compared methods (including fifteen existing representative Re-ID models, and two state-of-the-art Occluded Re-ID methods) are listed in Table~\ref{tab:single}. 
The results show that: 1) Compared to existing representative Re-ID models, our method achieves comparable performances with state-of-the-art on both datasets. 
These models leverage complicated attention mechanisms or local-based methods to achieve the results, while our PISNet is specifically designed for PI Re-ID. 
2) Our method outperforms the existing Occluded Re-ID models on both general Re-ID datasets.
Specifically, our method can reach 95.6\% rank-1 accuracy and 87.1\% mAP on Market1501, and 88.8\% rank-1 accuracy and 78.7\% mAP on DukeMTMC-ReID. 
The results prove the generalization ability of PISNet on the general Re-ID datasets.

\noindent \textbf{Influence of Parameters.} We evaluate two key parameters in our modelling, the loss weights $\alpha$ and $\beta$ in Eq.(7). 
The two parameters would influence the performance of the proposed method. 
As shown in Figure~\ref{fig:sa}, when $\alpha$ and $\beta$ are set between 0.3 and 1.0, and 0.4 and 1.0, respectively, the performance does not change dramatically, which indicates that PISNet is not sensitive to the $\alpha$ and $\beta$ in the value ranges.

\subsection{Attention Visualisation}

\begin{figure*}
\begin{center}

\includegraphics[width=8cm]{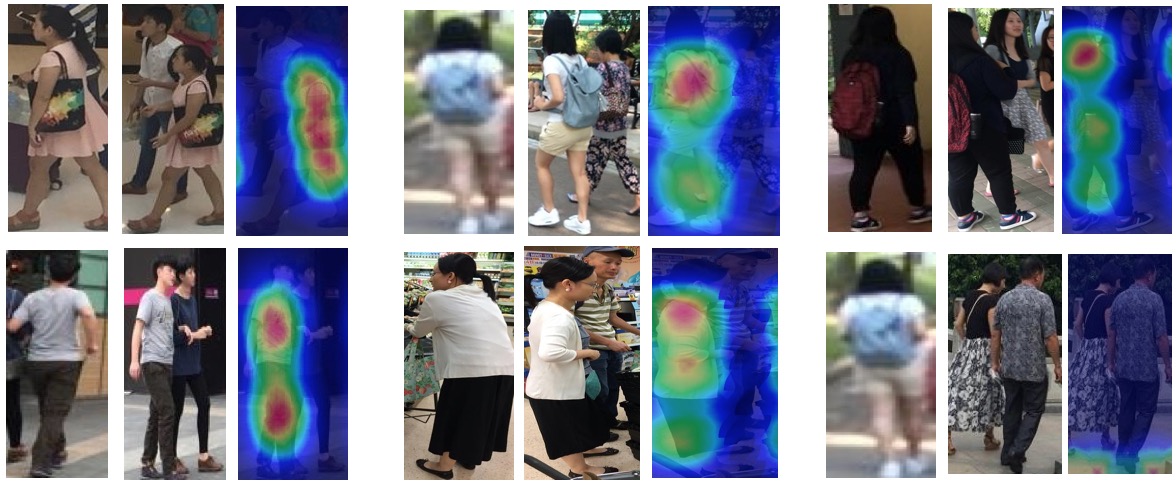}
   \caption{Visualisation of our query-guided attention for multi-person images in PI-CUHK-SYSU. 
   In each group, from left to right, (1) the single-person query, (2) the
   multi-person gallery and (3) the masked feature map.
   In the heat map, the response increases from blue to red. Best viewed in color
   }
 \label{fig:vis}
\end{center}

\end{figure*}

We visualise our query-guided attention for multi-person images in the PI-CUHK-SYSU dataset. 
Figure~\ref{fig:vis} shows that: (1) The attention mask filters out other pedestrians in multi-person images, 
(2) When the multi-person gallery does not include the query, the attention is weak for the whole image (the third group in the second rows).
The visualisation results further prove that our method can suppress the pedestrian interference effectively.


\section{Conclusions}

We have considered a new and more realistic person Re-ID challenge: pedestrian-interference person re-identification problem. 
To address the particular challenges associated with this new Re-ID problem, we propose a novel query-guided framework PISNet with a Guidance Reversed Attention Module and the Multi-Person Separation Loss. 
Both are specifically designed to address the person interference problem.
The effectiveness of our model has been demonstrated by extensive experiments on two new pedestrian-interference Re-ID datasets introduced in this paper.
In our future work, we will extend this work to handle more kinds of hard cases caused by a non-perfect detector.

\section*{Acknowledgment}
This work was supported by the National Key R\&D Program of China No. 2018YFB1004602 and the Project of the National Natural Science Foundation of China No. 61876210.

\clearpage
\bibliographystyle{splncs04}
\bibliography{egbib}
\end{document}